\documentclass[sigconf]{acmart}

\begin{document}

\title{MOROCCO: Model Resource Comparison Framework}

\author{Valentin Malykh}
\email{valentin.malykh@huawei.com}
\author{Ekaterina Artemova}
\email{Artemova.Ekaterina@huawei.com}
\affiliation{
  \institution{Huawei  Noah’s Ark lab}
  \city{Moscow}
  \country{Russia}
}

\author{Alexander Kukushkin}
\email{alex@alexkuk.ru}
\affiliation{
  \institution{Alexander Kukushkin}
  \institution{Data Science Laboratory}
  \city{Moscow}
  \country{Russia}
}

\author{Tatiana Shavrina}
\email{Shavrina.T.O@sberbank.ru}
\author{Vladislav Mikhailov}
\email{Mikhaylov.V.Nikola@sberbank.ru}
\author{Maria Tikhonova}
\email{m_tikhonova94@mail.ru}
\affiliation{
  \institution{Sberbank}
  \city{Moscow}
  \country{Russia}
}

\begin{abstract}
    The new generation of pre-trained NLP models push the SOTA to the new limits, but at the cost of computational resources, to the point that their use in real production environments is often prohibitively expensive. We tackle this problem by evaluating not only the standard quality metrics on downstream tasks but also the memory footprint and inference time. We present MOROCCO, a framework to compare language models compatible with \texttt{jiant} environment which supports over 50 NLU tasks, including SuperGLUE benchmark and multiple probing suites.  We demonstrate its applicability for two GLUE-like suites in different languages.\footnote{\url{https://github.com/RussianNLP/MOROCCO}}
\end{abstract}




\maketitle

\section{Introduction}
% Valentin

A new paradigm in natural language processing (NLP) has emerged in recent years. At the core of this paradigm is the notion of a pre-trained language model. Such models are usually pre-trained on a large number of unannotated texts using unsupervised objectives and only next fine-tuned for downstream tasks in a supervised or semi-supervised fashion. Pre-trained language models can be used either as sentence embeddings by producing vector representation for an input text or a sequence of vector representations for the text tokens. The expressive power of pre-trained language models allows establishing new state-of-the-art solutions for the majority of existing tasks, such as text classification~\cite{sun2019fine},  part-of-speech tagging (POS tagging)~\cite{tsai2019small}, machine translation~\cite{zhu2019incorporating}, and other. Simultaneously, the pre-trained language models require significant amounts of processing power, as they are mostly built from transformer blocks and number millions of parameters. 

Several benchmarks allow drawing a comparison between various language models in terms of the solution quality for downstream tasks or their capability to express linguistic information. To the best of our knowledge, none of the existing benchmarks account for computational efficiency and such characteristics as the memory fingerprint and the inference time of a language model at the same time. 
To this end, we propose a new evaluation methodology, which is aimed at measuring both model's performance and computational efficiency in downstream tasks.  Thus we introduce a novel MOdel ResOurCe COmparison framework (MOROCCO). We also provide a testbed for model evaluation in a fixed environment. 
Both the methodology and testbed are discussed in Section~\ref{sec:framework}. The evaluation results of the models on GLUE-like benchmarks and discussion on the methodology design are presented in Section~\ref{sec:results}.

\subsection{Related Work}
\label{sec:backgroud}
% Katya

\paragraph{NLP benchmarks}
Recently, multiple benchmarks aimed at natural language understanding (NLU) tasks have been established. The most prominent ones, GLUE~\cite{wang2018glue} and SuperGLUE~\cite{wang2019superglue}, set the trend for model-agnostic evaluation format. These benchmarks provide evaluation datasets and a public leaderboard. A submission to the leaderboard consists of predictions made on publicly available test sets. Thus, any model-specific parameters are not taken into consideration intentionally. GLUE and SuperGLUE benchmarks do not support any form of interaction with the model used to prepare the submission. The benchmarks offer nine general-domain NLU tasks in English. The more recent benchmarks follow same evaluation procedure but aim at domain-specific areas, such as dialogue systems \cite{MehriDialoGLUE2020} and biomedical NLU and reasoning \cite{gu2020domain}, or in the cross-lingual setting \cite{liang2020xglue,hu2020xtreme}.
Finally, at the beginning of 2021, a BERT-like model, DeBERTa~\cite{he2020deberta}, surpassed human performance on the SuperGLUE benchmark. This remarkable breakthrough was achieved with an architecture consisting of 48 Transformer layers counting 1.5 billion parameters. However, the comparison of DeBERTa's computational efficiency with other less performative models is left outside the SuperGLUE leaderboard.  

\paragraph{Efficient NLP}
% The vast majority of academic benchmarks do not account for computational efficiency. Memory footprint and inference time of language models are ignored or considered unimportant. However, in industrial applications computational efficiency is as important as model accuracy. 
The trade-off between model performance and computational efficiency has been explored in multiple shared tasks and competitions. The series of Efficient Neural Machine Translation challenges~\cite{birch2018findings,hayashi2019findings,heafield2020findings} measured machine translation inference performance on CPUs and GPUs with standardized training data and hardware. The performance was evaluated by the BLUE score, while the computational efficiency was measured by multiple parameters, including real-time, which the model used to translate the private test set, peak RAM and GPU RAM consumption, size of the model on disk, and the total size of the docker image, which could have included rule-based and hard-coded approaches. The organizers did not set up any restrictions on the measured parameters. Finally, the organizers selected the Pareto-optimal submission, i.e., those that need less computational resources, delivering though comparable to other systems high quality.

The EfficientQA challenge~\cite{min2021neurips}  challenged the participants to create an effective NLP-system for a single task, the open-domain question answering. The competition committee, however, limited the submissions to a few different restrictions based on the Docker-container size: participants could compete in creating the most accurate self-contained QA-system under 6Gb, or under 500Mb, or in training the smallest system that achieves 25\% accuracy, or, finally, in building the most accurate question answering system, regardless of size. Such restrictions have drawn the community's attention to studying the trade-off between storing the parameters of the pre-trained models + retrieval data or making smaller systems with model compression techniques + less redundant data.

The SustaiNLP challenge~\cite{wang2020overview} was aimed at measuring inference on the SuperGLUE benchmark. The efficiency is estimated by the power consumed throughout the course of inference. Submitted systems were run on standardized hardware environments. Experiment impact tracker  \cite{henderson2020systematic} measures energy consumption in kWh for submitted systems. The submitted systems improve total energy consumption over the BERT-base as much as $20\times$, but the results on average around 2 absolute points lower. The goal of the SustaiNLP challenge was to develop efficient but yet accurate models. Although using the same testbed, the MOROCCO framework was developed with the opposite goal in mind. It provides adequate estimates of how many resources consume the models that reach human-level performance. The MOROCCO framework aims at the latter. As MOROCCO supports Docker images, it can be easily integrated into any benchmark or probing task, built upon \texttt{jiant} framework\footnote{\url{https://github.com/nyu-mll/jiant}} described in~\cite{pruksachatkun2020jiant}.  

% \paragraph{Efficient Models} 
% Reformer https://arxiv.org/pdf/2001.04451.pdf

\section{Evaluation Framework}
\label{sec:framework}
% Valentin, Sasha
We present framework for the evaluation and a testbed\footnote{We will provide the links to the testbed website and the framework source code once the review process is over.}, where we guarantee the compatibility of achieved results. For the testbed a person (or a team) should prepare their submission as an Docker container and send it to the testbed. The testbed platform runs the solution Docker container with limited memory, CPU/GPU, and running time. The container is expected to read the texts from the standard input channel and output the answers to standard output. During the inference, the running time is recorded and later used for the submission scoring. To eliminate the running time and memory footprint dispersion caused by technical reasons, we  perform several runs and compute the median values.
The output of the container is evaluated with the task-specific metric. The resulting metric values are then used to compute the final evaluation score for the whole submission. To ensure the comparability of the collected metrics, we fix the hardware used for the computation. We use Yandex.Cloud\footnote{\url{https://cloud.yandex.com/}} virtual instances, where the following hardware is guaranteed: 1 $\times$ Intel Broadwell CPU, 1 $\times$ NVIDIA Tesla V100 GPU. The Docker container OS we use is Ubuntu 20.04.
Our framework is designed to comprise with \texttt{jiant} framework, alongside with simple requirements for the evaluation containers built upon other frameworks, and can be run locally avoiding our testbed usage.

\subsection{Metrics}
\label{sec:metrics}
As previous works mostly consider only the quality of the solutions, there are two important characteristics, namely, memory footprint and inference speed (throughput), which reflect the computational efficiency of a model.

\textit{Memory footprint}: to measure model GPU RAM usage $M$ we run a container with a single record as input, measure maximum GPU RAM consumption, repeat the procedure 5 times and compute a median value.

\textit{Inference speed}: to measure throughput we run a container with $N$ records as input, with batch size 32 and measure $T_N$. On all tasks batch size 32 utilizes GPU almost at 100\%. We also estimate initialization time $T_{\text{init}}$ with running a container with an input of size 1. Inference speed (throughput) $Tp$ is computed as follows: $$Tp = \frac{N}{T_{N} - T_{\text{init}}}.$$ In our experiments we use $N=2000$. We repeat the procedure 5 times to compute a median value.

\begin{figure}
    \centering
    \includegraphics[width=\linewidth]{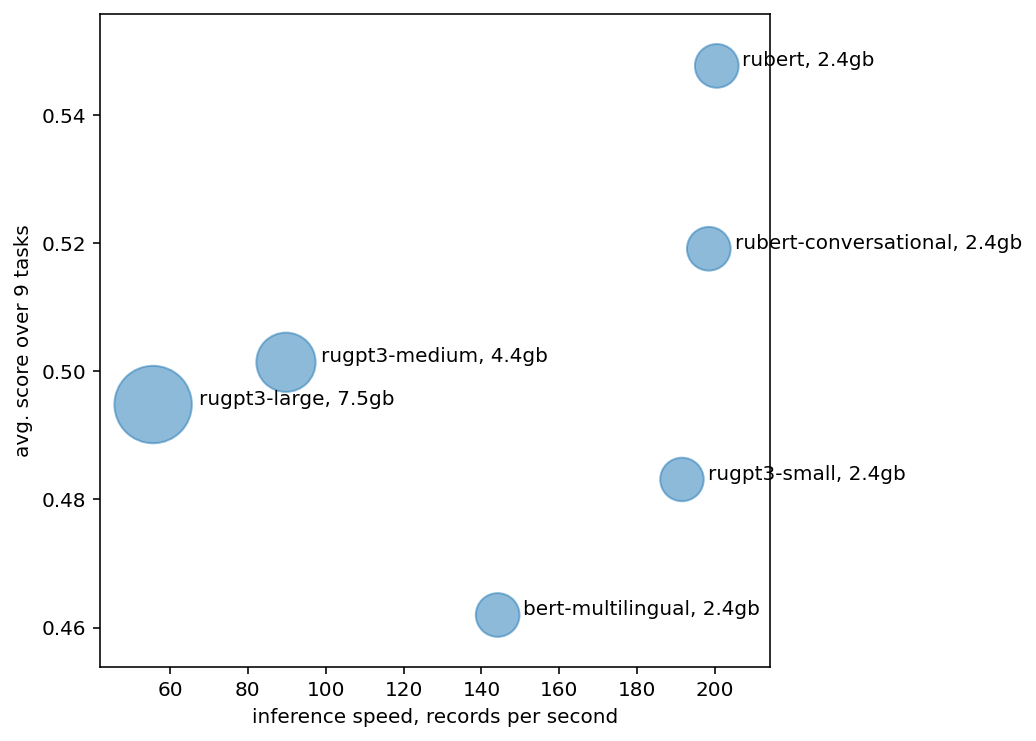}
    \caption{Models comparison on RussianSuperGLUE benchmark.}
    \label{fig:russianglue}
\end{figure}

We propose to use these three characteristics, namely $Q$, $Tp$, and $M$, in the following way: we comprise a 2-dimensional plot with  horizontal axis being a quality for a downstream task $Q$ (this metric is specific to the task) and vertical axis being a throughput $Tp$ for the model. To visualize memory footprint $M$ we propose to use circles of different sizes instead of a mere point on the plot. The example of such a plot is presented at Figure~\ref{fig:russianglue}.

We propose to take into account these three  characteristics of a model and make an integral measure of its ``fitness'' $F$ as follows:

$$F = Q \times \frac{Tp}{\log (M)},$$
where $Q$ is the metric-based measurement of the task-solving ability of a model, $M$ is measured in bytes, $Tp$ is measured in samples per second. We logarithm memory consumption $M$ since the model size increase is exponential for the modern models \cite{sanh2019distilbert}. This measure is motivated by the common idea that memory consumption should be lowered, while the achieved quality and processing speed should be increased. Thus it allows us to have the single value describing the efficacy of the model resource consumption. 

\subsection{Datasets}
In our work we run MOROCCO evaluation on SuperGLUE, and Russian SuperGLUE\footnote{The \url{https://russiansuperglue.com/}}~\cite{shavrina2020russiansuperglue} benchmarks for English and Russian respectively. The latter is a Russian counterpart of the English language SuperGLUE benchmark. Its tasks are organized analogously to the SuperGLUE. Namely, both benchmarks comprise 9 downstream tasks\footnote{SuperGLUE benchmark also includes additional Winogender Schema Diagnostics task which is a dataset designed to test for the presence of gender bias in automated coreference resolution systems. However, as long as it is not included in Russian SuperGLUE  we did not run MOROCCO evaluation on it.}: %, which could be divided into 5 groups according to the model skills they test: 
     \textit{Recognizing Textual Entailment} task is aimed to capture textual entailment in a binary classification form; %Given two text fragments (premise and hypothesis), the task is to determine, whether the meaning of hypothesis is entailed from the premise. 
     \textit{Commitment Bank}, which belongs to the natural language inference (NLI) group of tasks type with classification into 3 classes (entailment, contradiction, and neutral);
     \textit{Diagnostic dataset}, which is in fact another test set for recognizing textual entailment task additionally supplied with vast linguistic and semantic annotation;
     \textit{Words in Context} task is based on word sense disambiguation problem in a binary classification form;
     \textit{Choice of Plausible Alternatives} is a binary classification problem, which is aimed at accessing commonsense causal reasoning;
     \textit{Yes/No Questions} is a question answering task for closed (binary) questions;
     \textit{Multi-Sentence Reading Comprehension} is a task on machine reading, where the goal is to choose the correct answers for the questions based on the text paragraph;
     \textit{Reading Comprehension with Commonsense Reasoning} is a task on machine reading, where it is required to fill in the masked gaps in the sentence with the entities from the given text paragraph;
 \textit{Winograd Schema Challenge} is devoted to co-reference resolution in a binary classification form.
Aggregated information about the tasks is presented in Table~\ref{table:tasks}.
\begin{table*}[th!]
\begin{tabular}{|p{4cm}|l|l|l|l|l|l|}
\hline
\multicolumn{1}{|c|}{{\textbf{Task}}}                                        & \multicolumn{1}{c|}{{\textbf{Task Type}}} & \multicolumn{2}{c|}{\textbf{SuperGLUE}}                                           & \multicolumn{2}{c|}{\textbf{Russian SuperGLUE}}                                   & \multicolumn{1}{c|}{{\textbf{Task Metric}}} \\ \cline{3-6}
\multicolumn{1}{|c|}{}                                                                      & \multicolumn{1}{c|}{}                                    & \multicolumn{1}{l|}{\textbf{Name}} & \multicolumn{1}{l|}{\textbf{Samples}} & \multicolumn{1}{l|}{\textbf{Name}} & \multicolumn{1}{l|}{\textbf{Samples}} & \multicolumn{1}{c|}{}                                      \\ \hline
Recognizing Textual Entailment                                                                              & NLI                                                      & RTE                                 & 2490/277/3000                                   & TERRa                                &  2616/307/3198                                  & Acc                                         \\\hline
Commitment Bank                                                                              & NLI                                                      & CB                                 & 250/56/250                                   & RCB                                & 438/220/438                                  & Avg. F1 / Acc                                         \\\hline

Diagnostic                    & NLI \& diagnostics                                       & AX-b                               & 0/0/1104                                     & LiDiRus                            & 0/0/1104                                     & MCC                                                        \\\hline
Words in Context                                                                            & Common Sence                                             & WiC                                & 5428/638/1400                                & RUSSE                              & 19845/8508/18892                             & Acc                                                   \\\hline
Choice of Plausible  Alternatives                & Common Sence                                             & COPA                               & 400/100/500                                  & PARus                              & 400/100/500                                  & Acc                                                   \\\hline
Yes/No Questions                                                                                     & World Knowledge                                          & BoolQ                              & 9427/3270/3245                               & DaNetQA                            & 1749/821/805                                 & Acc                                                   \\\hline
Multi-Sentence Reading  Comprehension            & Machine Reading                                          & MultiRC                            & 456/83/166                                   & MuSeRC                             & 500/100/322                                  & F1 / EM                                                    \\\hline
Reading Comprehension with  Commonsense Reasoning & Machine Reading                                          & ReCoRD                             & 65709/7481/7484                              & RuCoS                              & 72193/7577/7257                              & F1 / EM                                                    \\\hline
The Winograd Schema  Challenge                    & Reasoning                                                & WSC                                & 554/104/146                                  & RWSD                               & 606/204/154                                  & Acc                                           \\\hline       
\end{tabular}
\caption{Datasets statistics. MCC stands for Matthews Correlation Coefficient; Acc - Accuracy; EM - Exact Match. The size train/validation/test splits are  provided in ``Samples'' columns.} \label{table:tasks}
\end{table*}

\subsection{Models}
\label{sec:models}
We run the experiments on the following publicly available models that achieved competitive performance on both SuperGLUE and Russian SuperGLUE benchmarks.
 \textit{Models for English} include monolingual (en\_bert\_base) and multilingual BERT (bert) \cite{devlin2018bert}, both in ``base'' variant, RoBERTa \cite{liu2019roberta} in ``base'' variant (en\_roberta\_base), ALBERT \cite{lan2019albert} in ``base'' variant (albert), and GPT-2  in ``large'' variant \cite{radford2019language} (en\_gpt2).
\textit{Models for Russian} involve multilingual BERT in ``base'' variant (bert-multilingual), 3 variants of ruGPT-3\footnote{\url{https://github.com/sberbank-ai/ru-gpts}} (rugpt3-small, rugpt3-medium, and rugpt3-large), Russian BERT (rubert) \cite{kuratov2019adaptation} in ``base'' variant, and its derived version Conversational RuBERT\footnote{\url{https://huggingface.co/DeepPavlov/rubert-base-cased-conversational}} in ``base'' variant (rubert-conversational).
All of the models  are released as a part of HuggingFace Transformers framework described in~\cite{Wolf2019HuggingFacesTS}.

\begin{figure}
    \centering
    \includegraphics[width=\linewidth]{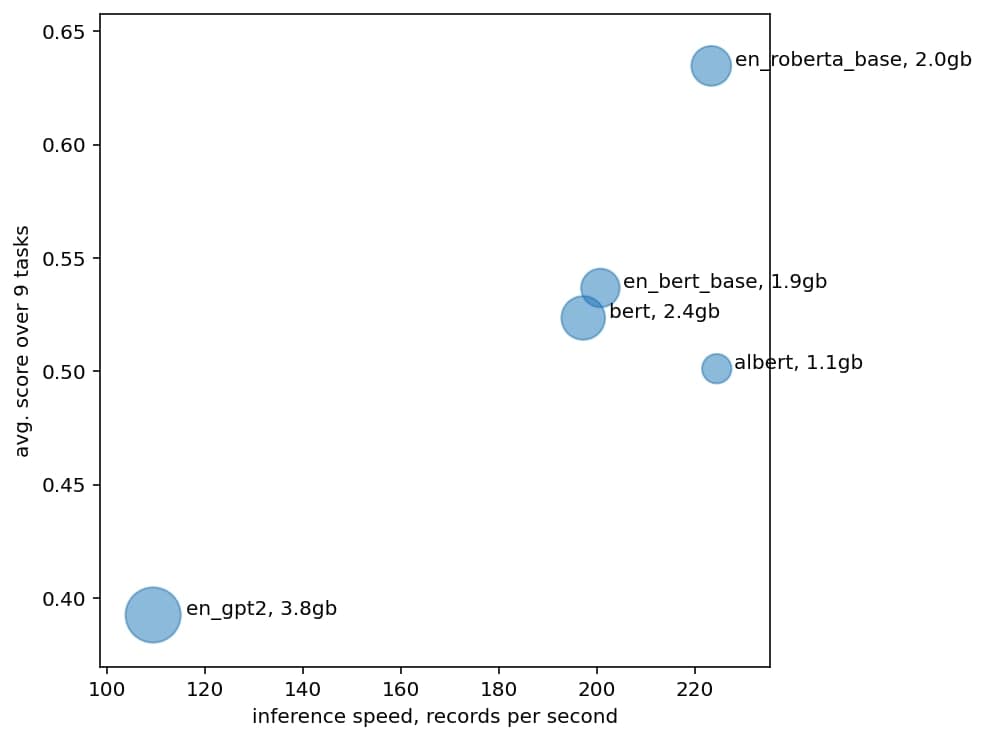}
    \caption{Models comparison on SuperGLUE benchmark (English).}
    \label{fig:englue}
\end{figure}

\section{Results}
\label{sec:results}
% Sasha, Vlad, Valentin 
We have measured $Q$, $Tp$, and $M$ for the models listed in the previous section and have drawn two figures demonstrating the results: Fig.~\ref{fig:englue} for SuperGLUE evaluation and Fig.~\ref{fig:russianglue} for RussianSuperGLUE one. 
We also evaluated $F$, the results are presented in Tab.~\ref{tab:fitness}. As one could see RoBERTa model had shown the best fitness for English language, while RuBERT is the best fit for Russian among the tested models. Overall these evaluations allowed us to separate better models by the means of quality, memory footprint, and throughput from the models showing worse performance and greater resource consumption.

\begin{table}[th!]
\small
\begin{center}%[htbp]
\begin{tabular}{|lllll|llllll|}\hline
\multicolumn{5}{|c|}{English} & \multicolumn{6}{c|}{Russian}\\
\rotatebox{90}{en\_bert\_base} & \rotatebox{90}{bert} & \rotatebox{90}{en\_roberta\_base}    & \rotatebox{90}{albert} & \rotatebox{90}{en\_gpt2} 
& \rotatebox{90}{bert-multilingual} & \rotatebox{90}{rugpt3-small} & \rotatebox{90}{rugpt3-medium}    & \rotatebox{90}{rugpt3-large} & \rotatebox{90}{rubert} & \rotatebox{90}{rubert-conversational} \\\hline
5.05 & 4.79 & \textbf{6.63} & 5.41 & 1.95 & 3.30 & 3.89 & 1.89& 1.24 & \textbf{4.84} & 4.59 \\\hline
\end{tabular}
\caption{Fitness evaluation for the models in two languages.}
\label{tab:fitness}
\end{center} 
\end{table}

\subsection{Discussion}
Our methodology has some limitations: we use averaging to estimate the values of $Q$, $Tp$, and $M$. While $M$ computation is least questionable, since the memory consumption for a single sample is more or less stable for any reasonable sample size, the other two measures require more attention. 

\begin{figure}
    \centering
    \includegraphics[width=\linewidth]{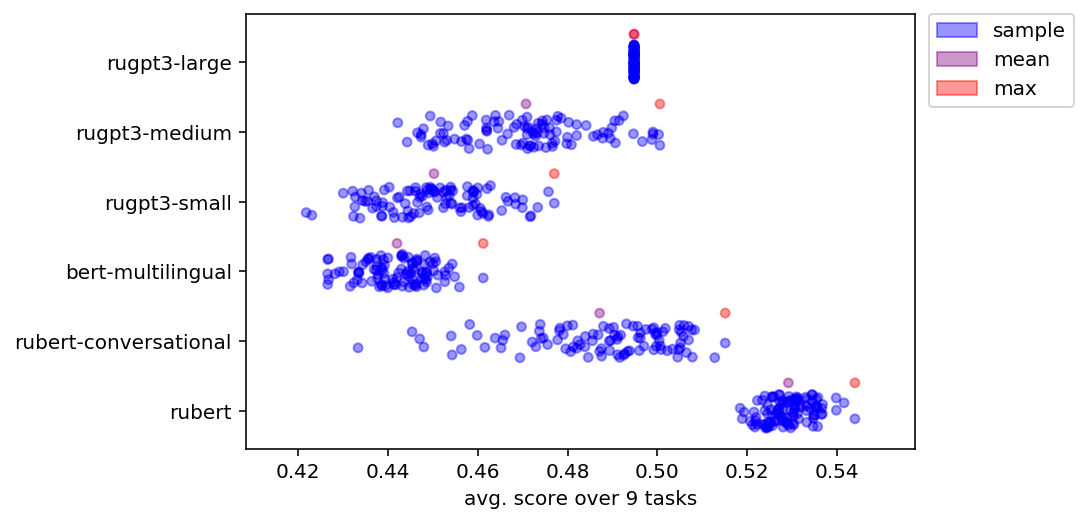}
    \caption{Quality comparison for mean and best results.}
    \label{fig:quality}
\end{figure}

We compare the mean and maximal quality values, as the latter is used on most of the leaderboards. The results of different models comparison on RussianSuperGLUE are presented in Fig.~\ref{fig:quality}. We show run results for ten evaluations for each of five different initializations of each model, with an exception for rugpt3-large, where we used only one initialization.\footnote{We add small random noise in vertical axis for better readability.} The ordering of the best and mean scores is keeping the same for mean (pale red) and maximal (full red) results, again with an exception for  rugpt3-large.

\begin{figure}
    \centering
    \includegraphics[width=\linewidth]{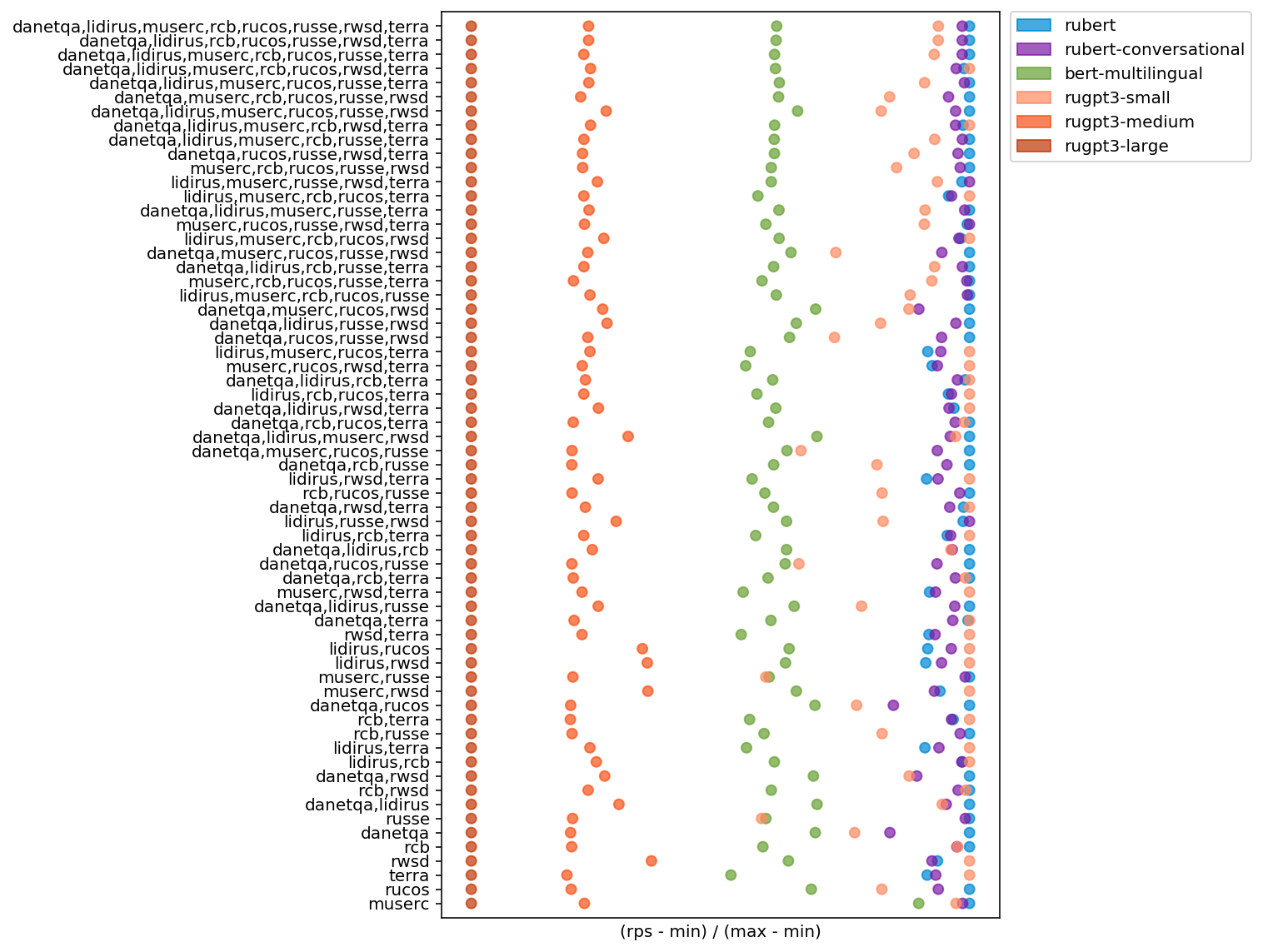}
    \caption{Throughput comparison for different datasets scores being averaged.}
    \label{fig:throughput}
\end{figure}

Another evaluation is presented at Fig.~\ref{fig:throughput}. We compare different sets used for averaging in RussianSuperGLUE by the synthetic value of normalized throughput. The normalization is done alongside the horizontal axis, thus one can compare the ordering for the models in different task sets. As one could see, the ordering is mostly keeping the same, with some occasional switches between the top models.

Based on this additional evaluation, we suppose that our methodology is stable regarding the choice between averaging schemes and the deviation in the max quality estimation process while being informative for the model comparison.

\section{Conclusion}
% Tanya
In this work, we presented the MOROCCO framework, which allows comparing NLP-models not only based on their overall quality metrics, but also on their resource consumption: the memory footprint and inference time. The proposed fitness metric (see section~\ref{sec:metrics}) allows us to compose the model leaderboard in a new way: to order them so that the most high-precision, smallest and fastest models are in the top, the accurate ones, but bigger and slower models are in the middle, and the most imprecise, largest and slowest ones are at the very bottom.
Thus, to obtain a higher place on the leaderboard researchers need to strive not for accuracy per cent fractions on the individual tasks, but for an overall improvement in both the performance and size of the model. A similar conditional assessment of the results has been adopted in computer vision, and since the last year in question answering.

The presented framework is compatible with the \texttt{jiant} framework and transformer models, making it easily applicable to evaluate a wide range of popular architectures, both multilingual and monolingual.

We hope that our work will initiate a more intensive search for a compromise evaluation of the overall performance of NLP-models, which could be an alternative to the existing dominant ``bigger is better'' methodology and would take into account the problems of overfitting, over-parametrization, data redundancy, and others.

As part of future work, we are considering closer cooperation with NLP-developers and enthusiasts to further search for the best industrial solutions, including organizing the competition of multilingual NLP-models on existing benchmarks as a possible step.

\bibliographystyle{ACM-Reference-Format}
\bibliography{lit}

\end{document}